\pdfoutput=1

\documentclass[11pt]{article}

\usepackage{ACL2023}

\usepackage{times}
\usepackage{latexsym}

\usepackage[T1]{fontenc}

\usepackage{CJKutf8}

\usepackage{microtype}

\usepackage{inconsolata}

\usepackage{graphicx}
\usepackage{booktabs}
\usepackage{amsmath} 
\usepackage{amsfonts}
\usepackage{multirow}
\usepackage{makecell}
\usepackage{lipsum}

\title{Generative Input: Towards Next-Generation Input Methods Paradigm}

\author{
Keyu Ding$^\dag$$^\ddag$$^\S$,
Yongcan Wang$^*$$^\ddag$$^\S$,
Zihang Xu$^*$$^\ddag$$^\S$,
Zhenzhen Jia$^\ddag$$^\S$,
Shijin Wang$^\ddag$$^\S$,
Cong Liu$^\ddag$$^\S$,
Enhong Chen$^\dag$
\\
{$^\dag$University of Science and Technology of China} \\
{$^\ddag$State Key Laboratory of Cognitive Intelligence, iFLYTEK Research, China} \\
{$^\S$iFLYTEK AI Research} \\
$^\dag$\tt\{kyding,cheneh\}@ustc.edu.cn\\
$^\ddag$\tt\{kyding,ycwang12,zhxu13,sjwang3,congliu2\}@iflytek.com}

\begin{document}
\begin{CJK*}{UTF8}{gbsn}
\maketitle

\newcommand\blfootnote[1]{%
\begingroup
\renewcommand\thefootnote{}\footnote{#1}%
\addtocounter{footnote}{-1}%
\endgroup
}
\blfootnote{*Equal contributions.}

\begin{abstract}
Since the release of ChatGPT, generative models have achieved tremendous success and become the de facto approach for various NLP tasks. However, its application in the field of input methods remains under-explored. Many neural network approaches have been applied to the construction of Chinese input method engines(IMEs).Previous research often assumed that the input pinyin was correct and focused on Pinyin-to-character(P2C) task, which significantly falls short of meeting users' demands. Moreover, previous research could not leverage user feedback to optimize the model and provide personalized results.
In this study, we propose a novel \textbf{Gene}rative \textbf{Input} paradigm named \textbf{GeneInput}. It uses prompts to handle all input scenarios and other intelligent auxiliary input functions, optimizing the model with user feedback to deliver personalized results. 
The results demonstrate that we have achieved state-of-the-art performance for the first time in the Full-mode Key-sequence to Characters(FK2C) task. We propose a novel reward model training method that eliminates the need for additional manual annotations and the performance surpasses GPT-4 in tasks involving intelligent association and conversational assistance. Compared to traditional paradigms, GeneInput not only demonstrates superior performance but also exhibits enhanced robustness, scalability, and online learning capabilities.
\end{abstract}

\section{Introduction}
One of the primary objectives of IMEs is to assist users in efficient text input. In some Asian languages, such as Chinese, Japanese, and Thai, they do not use alphabetic characters and cannot be directly inputted through a standard keyboard. Users often need to employ commercial input software, such as Sogou Input Method\footnote{\url{https://pinyin.sogou.com/}}, iFlytek Input Method\footnote{\url{https://srf.xunfei.cn/}}, Google Input Method\footnote{\url{https://www.google.com/inputtools/services/features/input-method.html}}, and so on, to accomplish text input.

Pinyin serves as the official romanization system for the Chinese language. In China, there are two common keyboard input methods: the 9-key keyboard and the 26-key keyboard. Each of these keyboard inputs further includes different input modes. In practice, user input scenarios are highly complex, with typical input modes illustrated in Figure \ref{fig_various-pinyin}, which illustrates various potential input modes for the Chinese sentence “我爱自然语言处理”(I love natural language processing). Some of the possible input modes include:

\begin{figure}
\centering
\includegraphics[width=0.8\linewidth]{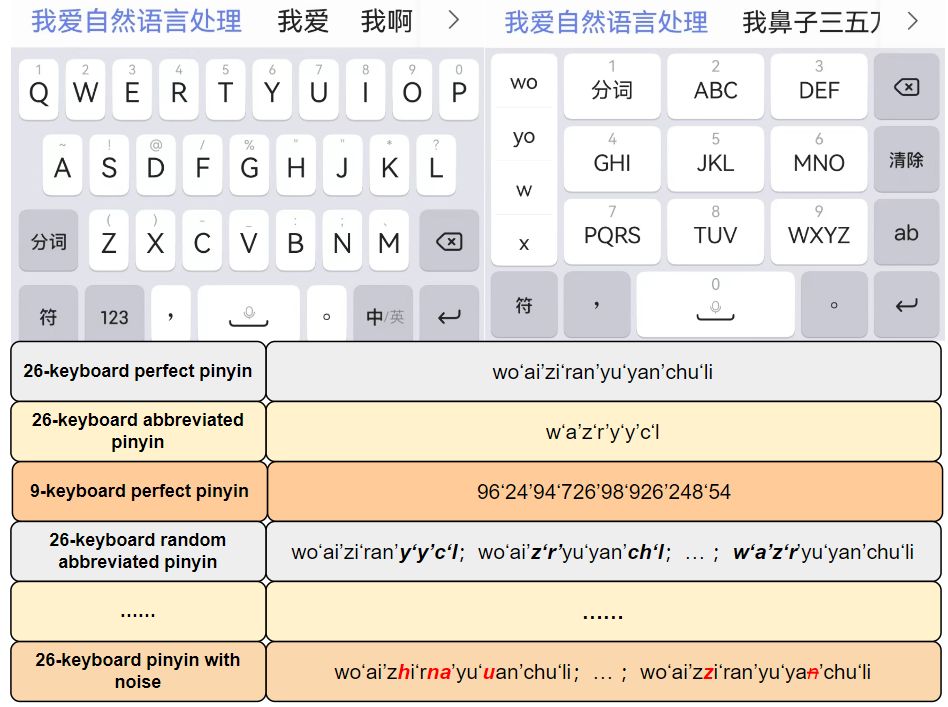}
\caption{\label{fig_various-pinyin}User PinYin input scenarios with typical input modes}
\end{figure}

1) 26-keyboard perfect Pinyin sequence(e.g., 'wo ai zi ran yu yan chu li'). 2) 26-keyboard abbreviated sequence(e.g., 'w a z r y y c l').
Both of these input modes have been extensively studied in prior works\cite{chenetal2015neural,tanetal2022exploring,xiao2022pert}. 3) 9-keyboard perfect Pinyin sequence (e.g., '96'24'94'726'98'926'248'54'). The 9-key keypad, as shown in the upper right corner of Figure 1, assigns 26 letters to 8 keys, with each key representing three to four pinyin characters, leading to a significant occurrence of homophones.
4) 26-keyboard random abbreviated pinyin sequence (e.g.,'wo ai zi ran y y c l'), which represents a mixture of perfect pinyin and abbreviated pinyin sequence. In this mixed scenario, there are numerous possible positions for the abbreviations to appear.
5) 26-keyboard pinyin with noise sequence (e.g., 'wo ai zhi rna yu uan chu li' ), which represents various input noise in the user's actual input process, including pinyin sequences or numeric sequences. Common errors include pressing an extra key (zi→zzi), missing a key (yan→ya), reversing the key order (ran→rna), and hitting the wrong key (yan→uan). In addition to errors related to keystroke actions, there are also errors stemming from dialectal differences, where some users may have difficulty distinguishing specific initials (声母) and finals (韵母). 

Apart from the listed typical cases, there are other types of noise, and these situations may occur randomly, collectively exhibiting an exponential growth pattern.To the best of our knowledge, there is currently no related research work covering such a wide range of practical input modes. Traditional input methods typically treat P2C as a sequence labeling task. In the initial stages, N-gram\cite{bahl1983maximum} models were used. In recent years, RNN models\cite{yao2018real,wu2017generating} have made significant progress in P2C tasks. Pre-trained models such as BERT-CRF\cite{souza2019portuguese} and GPT\cite{tan2022exploring}have also begun to be applied to sequence labeling tasks, such as named entity recognition and P2C conversion, and have significantly improved performance compared to RNNs.

\begin{figure}
\centering
\includegraphics[width=0.8\linewidth]{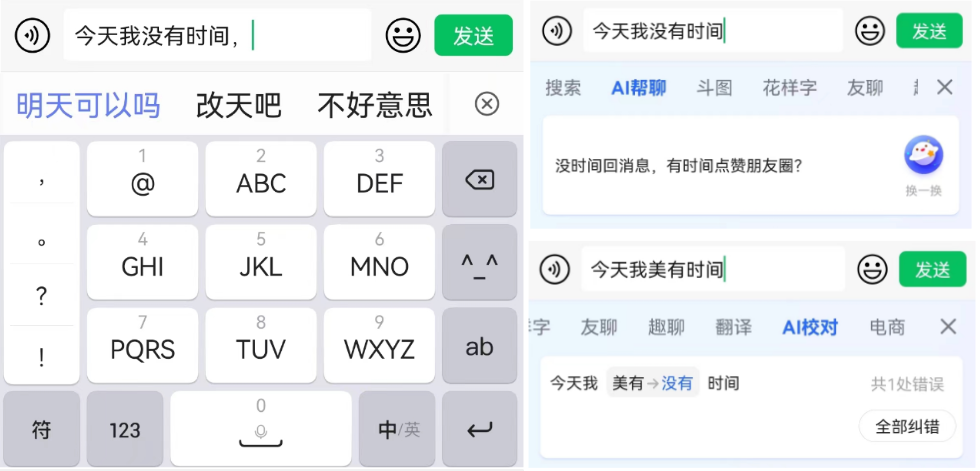}
\caption{\label{fig:various-function}Partial AI-assisted input scenarios}
\end{figure}

With the rapid development of AI technology, the functionality of input methods has far exceeded the P2C task. New features have emerged, such as intelligent association, conversational assistance, text correction,as Figure\ref{fig:various-function} shows, aimed at enhancing input efficiency, input enjoyment, and input accuracy. 

However, traditional paradigm-based IMEs have the following limitations,as Figure\ref{fig:tradition_and_geneInput}(a)shows:
\begin{itemize}
\item It is challenging to model noise in Pinyin input, as previous models often assume correct Pinyin input,
\item Assisting functions like intelligent association, conversational assistance, text correction are not unified in modeling, making the input system complex and fragile,
\item Cannot effectively utilize users' feedback for online optimization and cannot generate personalized results.
\end{itemize}

Driven by the flourishing development of AI-Generated Content (AIGC), exemplified by models like InstructGPT\cite{ouyang2022training} and GPT-4\cite{openai2023gpt4} that exhibit human-level content generation capabilities, IMEs have the opportunity to break free from the constraints of traditional paradigms. It becomes possible to model various input tasks in a unified text generation framework. In light of this, we propose a new paradigm for generative input methods, which offers the following advantages:

\begin{itemize}
\item It takes various input sequences, including those with noisy Pinyin input, as model inputs, covering all possible scenarios,
\item It unifies all kinds of tasks into a text generation task, utilizing a single model to handle all tasks, resulting in a highly robust system,
\item It employs reinforcement learning and Contrastive learning to learn from user feedback, automatically adjusting and optimizing the model and obtain adaptive results.
\end{itemize}





\section{Tasks}\label{sec:task}


    
    

We select the most representative three tasks in the input method as research objects.

\textbf{Full-mode Key-sequence to Characters (FK2C)}: This task is the core of input method, which converts the sequence of user keystrokes into Characters. Previous works usually assume that the user input is preprocessed Pinyin form, and directly model Pinyin to Characters conversion, so it is called P2C in short. But the actual input is some 26-key letter keystroke sequences or 9-key number keystroke sequences, and the user input sequence may not correspond to the complete Pinyin, and contains noise, the same key sequence may correspond to results of different input modes, as shown in Figure \ref{fig_various-pinyin}. Therefore, directly modeling the keystroke sequence to characters conversion is a more challenging task than the traditional P2C task.

\textbf{Intelligent Association (IntelAssoc)}: This task is a commonly used input assistance function, which predicts possible next sentences based on the content already entered by the user for selection, to improve input efficiency. It is a typical text continuation function, which mainly represents the generation of variable-length text scenarios.

\textbf{Conversational Assistance(ConvAssist)}: This task mainly involves text beautification of user content, rewriting the input content to meet specific requirements without changing the original semantics of the user's input, such as being more humorous or witty. It represents scenarios where the length of the input text is roughly equivalent to the output text.


Current related research on the input method mainly explores a single functional point. \citet{2018KNPTC} propose KNPTC to integrate letter-neighbor knowledge into NMT for Pinyin Error Correction. \citet{2018Open} followed users’ input behavior through an online updated lexicon. \citet{2018Moon} integrated attention-based neural machine translation (NMT) models and information retrieval (IR) into Pinyin input methods, providing interesting and customizable association capabilities. \citet{2018Chinese} encoded previous input sentences as additional context for learning, predicting character sequences of incomplete Pinyin inputs. \citet{tanetal2022exploring} further applied Chinese GPT to Pinyin input methods, solving the problem of incomplete Pinyin input effects by enriching Pinyin in the context. These works mainly focus on modeling a single input mode in the P2C task, lack attention to other tasks in the input method, and cannot meet the actual use requirements.

\section{Models}


\begin{figure}
  \centering
  \includegraphics[width=0.8\linewidth]{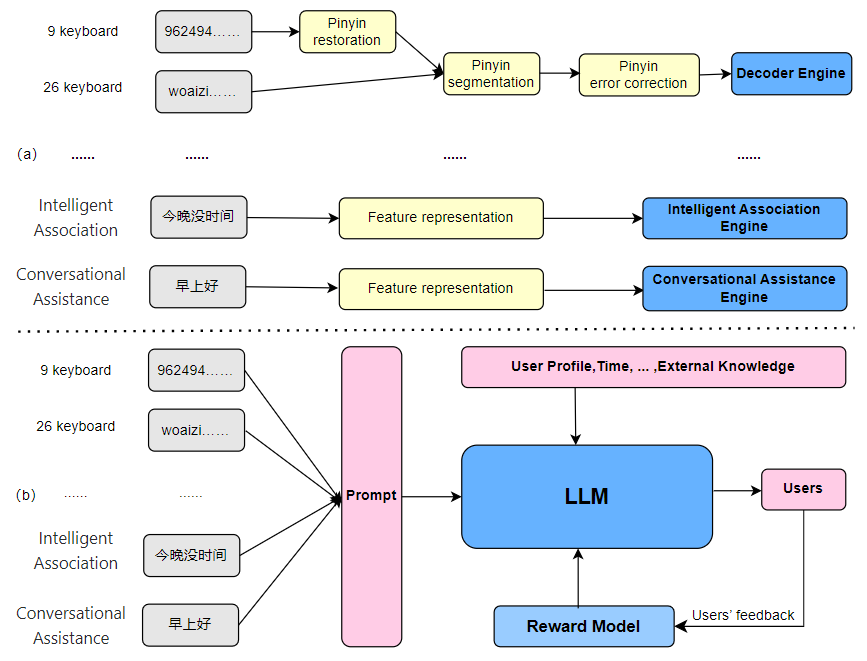}
  \caption{Comparison between the traditional(a) and GeneInput paradigm(b). }
  \label{fig:tradition_and_geneInput}
\end{figure}


 In this paper, we propose a generative modeling scheme GeneInput to uniformly model the typical tasks and different input modes contained in the input method scenario, as Figure\ref{fig:tradition_and_geneInput}(b) shows. It leverages an LLM to model Pinyin decoding tasks in various noisy scenarios, various AI-assisted input functionalities, and utilizes user feedback to automatically adjust and optimize the model. Additionally, it integrates historical user input information to provide personalized results.

\subsection{GeneInput}




Large language models (LLM) have achieved good results in many tasks, and studies show that LLM can distinguish tasks through different prompts, thus unifying the modeling of different tasks \cite{ouyang2022training} \cite{openai2023gpt4} \cite{2021Finetuned}. However, to our knowledge, there is no related work using LLM to uniformly model input method-related tasks, and the existing LLM are not ideal for input method-related tasks, GPT4 \cite{openai2023gpt4} also performs much worse than commercial input methods on the K2C task. Therefore, this work attempts to use LLM to uniformly model various typical input method tasks by setting different prompts.



To uniformly model the various functions of the input method scenario, we designed corresponding prompts for the three typical input method tasks introduced in section\ref{sec:task}, and fine-tuned them based on generative large language models. As shown in Figure\ref{fig:model}(a), in the model, given the corresponding task description P and input X, we predict the corresponding output character $Y=[y_{1}, ... ,y_{n}]$. The model training objective is to minimize the following loss function.

        \begin{equation}
        \small
        \begin{aligned}
        L=-\sum ^{n}_{j=1}\log p\left( y_{j}| y_{<j},P,X\right)
        \end{aligned}
        \end{equation}


Specifically, for the intelligent association task, the input X is the current input sentence and the output Y is the corresponding possible next sentence. For the conversation assistance task, the input X is the user’s original input sentence and the output Y is the beautified paraphrased sentence of this sentence. And for the K2C task, the input X is the user-input 26-key or 9-key keystroke sequence and the output Y is the corresponding Chinese character result. And for each task, we carefully design a corresponding task description P to ensure that the model knows the goal and requirements of each task, thus making clear distinctions among tasks. Since the full-mode K2C task is a more complex and challenging task, we have done more extended design for this task, which will be introduced in section\ref{sec:fk2c}.


The structure is simple and flexible, the task description, input, output in the model can be modified or extended according to different tasks. We can achieve more accurate prediction by extending the input, adding some additional information that can be obtained, such as content above or user information. It is also possible to add some important intermediate results in the output for more complex tasks, thereby reducing the difficulty of task modeling.

\subsection{Full-mode Key-sequence to Characters}\label{sec:fk2c}



The task of Full-mode Key-sequence to Characters, as a core function of the input method, has complex input modes. Previous work often significantly simplified this task, such as \citet{tanetal2022exploring} assuming that user input is always completely correct and pre-segmenting the input into pinyin, modeling only the single mode of perfect pinyin or abbreviate pinyin. However, in actual use, users' inputs are often noisy raw keystroke sequences, and it is impossible to predict whether users will input according to a certain deterministic mode, so considering all possible input modes, providing results for all reasonable input modes and ranking is crucial for the input method. To our knowledge, this paper is the first study to uniformly model the full input modes of the input method and directly act on the user's original input.



\begin{figure}
\centering
\includegraphics[width=0.8\linewidth]{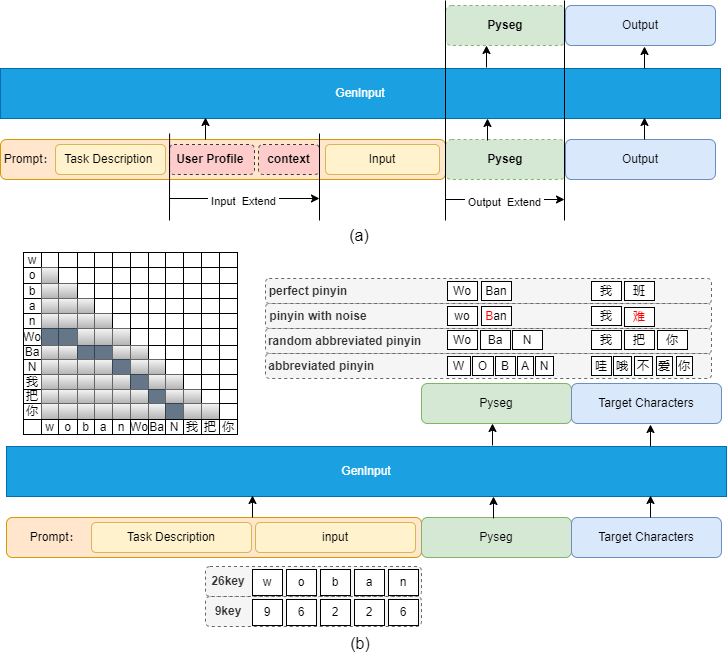}
\caption{\label{fig:model}The architecture of IME unified modeling (a) and full-mode K2C (b).}
\end{figure}

In general, the K2C process is usually divided into two stages. First, the keystroke sequence entered by the user is converted into the corresponding pinyin segmentation results, and then the corresponding text results are decoded according to different pinyin segmentations. Here, we will refer to the pinyin segmentation results as "pyseg" for simplicity. The same keystroke sequence can be segmented into different pinyin paths, and different segmentation paths generally correspond to different input modes. As shown in Fig\ref{fig:model}, when inputting the keystroke sequence "woban", it may be segmented into the perfect pinyin path "Wo'Ban", the abbreviate pinyin "W'O'B'A'N", or the random abbreviate pinyin "Wo'Ba'N". However, due to the mapping relationship of one digit keystroke to multiple letters, the situation is more complex for 9-key input. Therefore, as an important part of K2C, pinyin segmentation plays a significant role in the full-mode K2C of the input method.


So we incorporate the intermediate process of Pinyin segmentation in modeling, and conduct full-mode K2C modeling based on pyseg. And through pyseg to connect the input and output, impose alignment constraints, improve the quality of generated candidates.


\subsubsection{FK2C modeling based on pyseg}



Research shows that it is insufficient to directly model the mapping from input x to output y for complex problems, and the introduction of intermediate processes can greatly enhance the ability of LLM \cite{2022Chain}. The K2C task is different from open generation tasks such as intelligent association, which are strictly constrained by inputs and have relatively deterministic answers and objective evaluation criteria, and should have a relatively rigorous reasoning process. Therefore, in order to better unify the modeling of multiple modes, we add the Pinyin segmentation prediction task. As shown in Figure \ref{fig:model} (b), we add pyseg as an extension of the output, and based on pyseg, we uniformly model various input modes of the input method in all scenarios, enhancing the modeling ability of different input modes. First, from the task description P and the input keystroke sequence $X=[x_{1}, ... ,x_{m}]$, we predict the possible Pinyin segmentation $S=[s_{1}, ... ,s_{n}]$, and then combine the first two to predict the final result $Y=[y_{1}, ... ,y_{n}]$. After the output is extended, the corresponding training loss function is:

        \begin{equation}
        \small
        \begin{aligned}
        L_{pysegs}=-\sum ^{n}_{i=1}\log p\left( s_{i}| s_{<i},P,X\right)
        \end{aligned}
        \end{equation}

        \begin{equation}
        \small
        \begin{aligned}
        L_{words}=-\sum ^{n}_{j=1}\log p\left( y_{j}| y_{<j},S,P,X\right)
        \end{aligned}
        \end{equation}

        \begin{equation}
        \small
        \begin{aligned}
        L=\lambda \cdot L_{pysegs}+L_{words}
        \end{aligned}
        \end{equation}


In which, $\lambda$ is the hyperparameter and can be adjusted according to the importance of the expanded output part.










\begin{figure}
\centering
\includegraphics[width=\linewidth]{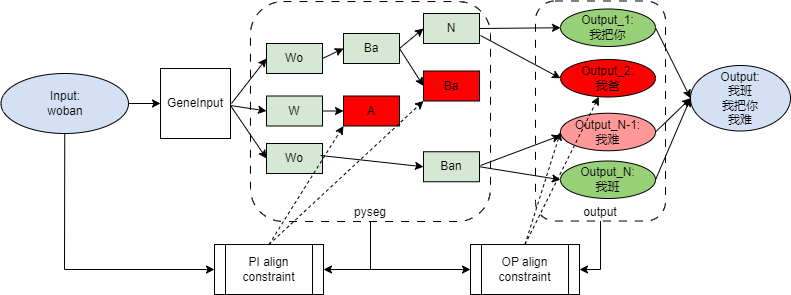}
\caption{\label{fig:decode}The full-mode k2C alignment constraint generation process based on pyseg}
\end{figure}

\subsubsection{Alignment Constraint Generation}









As shown in Figure \ref{fig_various-pinyin}, the prefect pinyin or abbreviated pinyin input corresponding to Chinese sentences is relatively singular, so there is a clear alignment relationship. However, the random abbreviated pinyin input will exponentially increase with the length of the characters, and the one-to-many mapping relationship in the 9 keys and the existence of input noise make the input situation corresponding to the sentence more complex. Therefore, So when full-mode modeling is performed, it is not easy to directly align each character in the output with the corresponding keystroke sequence of the input, especially when the input and output are long. However, the generated intermediate result pyseg can form a relatively clear alignment relationship with both input and output, so we propose an alignment constraint based on pyseg bridging. According to the generation process, the alignment constraint can be divided into two stages: pyseg-input alignment and output-pyseg alignment.



\textbf{Pyseg-Input Alignment} Since Pinyin segmentation only performs a limited spatial mapping and segmentation on the original input, each Pinyin segmentation result can be back-mapped to a uniquely determined input form and is consistent with the original input. Therefore, we can compare it with the original input by mapping the decoded Pinyin segmentation back to the corresponding 26-key or 9-key input, and remove the erroneous Pinyin segmentation paths that are inconsistent with the input. $s_{i}$ represents the generated i-th Pinyin, and its corresponding probability is as follows.


\begin{equation}
\small
\begin{aligned}
 p(s_{i})=\frac{exp(g(s_{i})) }{\sum_{P2I(s_{<j}s_{j}) \in \nu_{X} } exp(g(s_{j}))}
\end{aligned}
\end{equation}


In this context, g represents the logit before softmax, $\nu_{X}$ denotes the prefix subset corresponding to the original input X, and P2I() represents the mapping function from pinyin to input. That is, the current decoded pinyin segmentation path should be restored as a prefix subset of the input X, and the final complete pinyin segmentation path restored should be consistent with the input X. As shown in Figure 5, the middle path generates a Pinyin character "Ba" in the third step. After restoration, this Pinyin segmentation path becomes "wobaba", which is not in the prefix subset of the input "woban". Therefore, it can be known that this path is an illegal path.









\textbf{Output-Pyseg Alignment} Different from the uncertainty of alignment relationship between output string $Y=[y_{1}, ... ,y_{n}]$ directly corresponding to keystroke sequence $X=[x_{1}, ... ,x_{m}]$, the output string $Y=[y_{1}, ... ,y_{n}]$ and Pinyin segmentation $S=[s_{1}, ... ,s_{n}]$ correspond one-to-one. The Pinyin of each character $y_{i}$ should be consistent with the corresponding Pinyin segmentation $s_{i}$, otherwise there may be noise in the input, and the path can be penalized according to the inconsistent ratio. We impose a certain penalty on the results that are inconsistent with the Pinyin segmentation by comparing the edit distance between the generated Chinese characters and the corresponding position Pinyin segmentation, so as to avoid excessive error correction or generating completely unrelated results to the input, and the correction penalty coefficient for the i-th step is as follows.

        \begin{equation}
        \small
        \begin{aligned}
         \varepsilon_{i} =\frac{\alpha}{n}EditDist(s_{i},C2P(y_{i}, mode(s_{i}))) 
        \end{aligned}
        \end{equation}


In this context, n is the number of inputs corresponding to $s_{i}$, EditDistance(a, b) represents the edit distance between a and b. The C2P() function is used to romanize the generated Chinese characters $y_{i}$ and select the corresponding perfect pinyin or abbreviate pinyin based on the mode of $s_{i}$. $\alpha$ is a hyperparameter that adjusts the correction penalty strength.


After increasing the alignment constraints with Pinyin segmentation, $y_{i}$ represents the generated i-th Chinese character, and its corresponding probability is as follows.

    \begin{equation}
    \small    
    \begin{aligned}
     p(y_{i})=(1-\varepsilon_{i})\cdot \frac{exp(g(y_{i})) }{\sum exp(g(y_{j})}
    \end{aligned}
    \end{equation}


\subsection{IME Personalization}



For the same input, the expected results of different types of users often have significant differences. How to utilize available additional information to provide differentiated results and more accurately meet the needs of user inputs is key to enhancing the user experience of the input method.



With the user's informed consent and authorization, GenInput can conveniently incorporate existing historical input and user profile information to provide users with more accurate personalized results. Compared to complex encoding designs, GPT-like LLM can easily incorporate these additional pieces of information. We only need to add the previous context or detailed descriptions of known user-specific labels in the prompt as extended input information. The corresponding model training loss function is:

        \begin{equation}
        \small
        \begin{aligned}
        L=-\sum ^{n}_{j=1}\log p\left( y_{j}| y_{<j},P,E,X\right)
        \end{aligned}
        \end{equation}


In this context, E represents optional input expansion information. It can be the current input context information or user profile information, such as gender, age, occupation, hobbies, etc., or a list of high-frequency user inputs in history.

\subsection{Online Optimization with Human Feedback} \label{sec:imerlhfmethod}






Previously, the optimization of language models behind input method editors mainly followed the classical paradigm of ``pre-training + fine-tuning''\cite{gpt1}. However, such a complete model development process has very high requirements on the quality and quantity of training data, computational resources, time, and so on, so it is difficult to iterate models rapidly. In the input method scenario, the style and preference of people's daily communication language change fast with the passage of time, so the traditional paradigm can not meet the optimization needs of the input method models. Recently, the key technology behind ChatGPT, RLHF, can effectively help the model to follow the human preference\cite{ouyang2022training}. Therefore, we apply RLHF on fine-tuned large language models, so that its output on the downstream task is more in line with the requirements of the users of input method scenarios. We refer to this part as RLHF-IME (Reinforcement Learning from Human Feedback in Input Method Editor) in the following sections.


Studies show that the reward model is extremely important and its performance determines the upper bound of RLHF to some extent \cite{mossrlhf}. Since the reward model training data in \citet{ouyang2022training} requires a large number of high-quality trained annotators to participate in sorted annotation, which is too costly in terms of time and money, we design fully automated annotation methods that are more feasible and friendly to a large number of real-world application scenarios in the industry. Considering the text characteristics of the input method scenario, multiple reward model training methods based on two annotation systems are designed and put into use.

\begin{figure}
\centering
\includegraphics[width=0.8\linewidth]{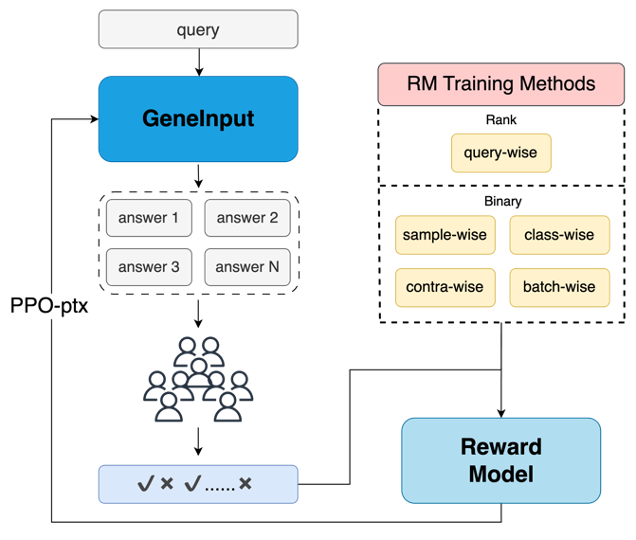}
\caption{\label{fig:rlhf2}Online optimization with human feedback framework}
\end{figure}

Figure \ref{fig:rlhf2} illustrates the whole process of online optimization with human feedback which consists of automatic data generation, reward model training, and GeneInput optimized with the reinforcement learning algorithm iteratively.

\subsubsection{Ranking System}

\citet{ouyang2022training} shows us that it is effective to train reward models based on manually labeled rankings of LLM outputs. For the purpose of labeling automatically, we believe that we can extract information fitting the preference of the whole user group from amounts of user behaviors of choosing answers, which can be used as the basis for ranking the answers. 


Taking three months as the statistical cycle, the labeling score of the answer is calculated based on the percentage of the number of times the answer is selected to the number of times the answer is provided to the users as a candidate answer as shown in Equation \ref{eq:ranklabel}, where $ N_{selected}^{i} $ is the time of $i_{th}$ sample been selected by users in the current statistical cycle.

\begin{equation}
\small
\begin{aligned}
label_{i} = \begin{cases}
0 \quad & if N_{selected}^{i} = 0, \\
max(1, \frac{N_{selected}^{i}}{N_{provided}^{i}}*100) \quad & otherwise.
\end{cases}
\label{eq:ranklabel}
\end{aligned}
\end{equation}

Under the ranking annotation system, we design the following reward model training method.




\textbf{Query-Wise} The Query-Wise method is conducted based on comparing samples with the same query but different answers. Under the same query, it is reasonable to rank different answers according to the user labeling scores and train models to judge the quality of answers, otherwise, comparing the scores is meaningless. The loss function is constructed by the formula \ref{eq:better_samples_num} and \ref{eq:querywiseloss}, where n denotes the number of answers of the current query, $query_i$ means the query of the $i_{th}$ sample and $label_i$ means the human preference score of the $i_{th}$ answer based on the fully automatic labeling scheme introduced above.

\begin{equation}
\small
\begin{aligned}
ld(i,j) = label_i - label_j
\label{eq:labeldiff}
\end{aligned}
\end{equation}

\begin{equation}
\small
\begin{aligned}
bt(i) = \sum_{j=1}^{n}\mathbb{I}_{query_i = query_j}\mathbb{I}_{ld(i,j) > 0}\cdot1
\label{eq:better_samples_num}
\end{aligned}
\end{equation}


\begin{equation}
\small
\begin{aligned}
\mathcal{L}_{\textrm{QW}} = -\frac{1}{n}\sum_{i=1}^{n} \frac{1}{bt(i)} \sum_{j=1}^{bt(i)}\ln_{}{\sigma[s(x, y_{i}) - s(x,y_{j})]^{ld(i,j)^{-1}}}
\label{eq:querywiseloss}
\end{aligned}
\end{equation}



\subsubsection{Binary Classification System}

Texts in input method scenarios are characterized by high contextual diversity and short length (less information in a single sentence), so it is challenging to rank different answers in one order because most of them can be correct in some specific contexts. For example, in the intelligent association task, when the query is ``I haven't slept yet'', the candidate answers  ``because I haven't finished my homework yet'', ``because I'm still working overtime'', ``because I drank too much coffee and am suffer from insomnia'' may be the correct or preferred one for different users. However, for answers that are necessarily impossible (e.g., when the answer is ``Steak and black pepper go well together''), people can often make a clear distinction. Therefore, given this situation, we believe that we can try to classify the answers into two classes, i.e., whether they are likely to be reasonable answers or not. In this system, we use a fully automated labeling scheme for three months, where answers that have been selected by real users during the period are labeled as positive answers and those that have not been selected are labeled as negative. Based on this, we design various reward model training methods as follows:


 \textbf{Sample-Wise} This is the sample-level training method based on each sample for a binary classification task, allowing the model to judge whether the current answer is correct or not given the specific task and query. The loss function is the binary cross-entropy loss as shown in equation \ref{eq:samplewiseloss}, where $y_{i}$ and $\hat{y_{i}}$ denote the true category of the current sample and the probability that the model predicts the correct category, respectively.

\begin{equation}
\small
\begin{aligned}
\mathcal{L}_{\textrm{SW}} = -\frac{1}{n}\sum_{i=1}^{n} [y_{i}·\log\hat{y_{i}}+(1-y_{i})·\log(1-\hat{y_{i}})] \label{eq:samplewiseloss}
\end{aligned}
\end{equation}


\textbf{Class-Wise} In the training method for category granularity, we expect models to acquire the ability to judge the correctness of an answer by taking the samples from the two categories as each as a whole and training the model so that its score for the correct answers is greater than that for the incorrect. The loss function is shown in Equation \ref{eq:classwiseloss}, where $n$ and $m$ denote the number of samples of correct and incorrect answers, respectively. $x$ stands for query, $s(x,y)$ denotes the reward model's scoring of answer $y$ when the query is $x$.

\begin{equation}
\small
\begin{aligned}
\mathcal{L}_{\textrm{CW}} = -\log{\sigma(\frac{1}{n}\sum_{i=1}^{n}s(x, y_{i}) - \frac{1}{m}\sum_{j=1}^{m}s(x, y_{j}))} \label{eq:classwiseloss}
\end{aligned}
\end{equation}


\textbf{Batch-Wise} Batch-Wise is a kind of pair-level training method. For the data within a batch, we pair every correct and incorrect answer, then calculate the difference between model scores to construct a pair-grained training loss. In this method, we do not require that the queries of the two samples in the pair are the same because, under the binary annotation system, we believe that the model scores of any correct answer should always be higher than any incorrect answer, no matter what the query is. For example, in the intelligent association task, the model should judge \texttt{[query="早上好(Good morning)", answer="吃早饭了吗？ (Have you had breakfast yet?)"]} to be superior to \texttt{[query="今天天气不错(it's a nice day)", answer="衬衫的价格是九榜十五便士(the price of the shirt is nine pounds fifteen pence)"]}. For the loss function, please refer to Equation \ref{eq:batchwiseloss}, where n and m refer to the number of correct and incorrect answers in a batch, respectively.

\begin{equation}
\small
\begin{aligned}
\mathcal{L}_{\textrm{BW}} = - \frac{1}{nm}\sum_{j=1}^{n} \sum_{j=1}^{m}\log_{}{\sigma[s(x, y_{i}) - s(x, y_{j})]}
\label{eq:batchwiseloss}
\end{aligned}
\end{equation}

\textbf{Contra-Wise} Contrastive learning has been verified to be effective in many areas of NLP\cite{simcse}, so we introduce supervised contrastive learning\cite{Khosla2020SupervisedCL} into reward model training by using the categorization information of the samples as supervised signals for the positive and negative examples, expecting the model to improve the ability to judge the strengths and weaknesses of the answers by means of comparing the positive and negative examples with a loss function constructed in line with \citet{Khosla2020SupervisedCL}.



In the reinforcement learning stage, we use the training method of the model ``ppo-ptx'' in \citet{ouyang2022training}, i.e., combining ppo loss with pre-train loss. During the training process, we use the reward models trained based on the above methods to score the answers generated by fine-tuned Spark. The reward models serve as an approximation of human preferences to guide the optimization direction of the generative large language models.

\section{Experiments}

\subsection{Experiments Settings}

\subsubsection{Public Datasets} 


\textbf{PD/TP Datasets} The PD dataset \cite{2012A} and TP dataset \cite{2017Tracing}  are currently publicly available datasets commonly used to evaluate the P2C effect of input methods. The PD dataset is extracted from the People's Daily corpus from 1992 to 1998. Meanwhile, The TP dataset is constructed from user chat logs collected by TouchPal IME. Each dataset contains 2000 test data points, but all only include perfect pinyin input.
 
\subsubsection{XF Datasets} \label{sec:xfdatasets}

Due to the lack of publicly available datasets for intelligent association (IA), conversational assistance (CA), and full-mode K2C in input methods, we constructed a new dataset for input method tasks called the XF dataset.

\textbf{SFT Datasets}  We separately built an IA dataset containing 8 million context pairs, a CA dataset with 6 million instances, and a full-mode K2C dataset with 12 million entries. Additionally, we constructed 1,000 intelligent association test sets and 1,000 conversational assistance test sets for manual evaluation of the final results. For the evaluation of FK2C task effects under different keyboards of 26-key and 9-key, we constructed a test set of 57k, covering different input modes such as perfect pinyin, abbreviated pinyin, random abbreviated pinyin, and noisy input with different error types.




\textbf{RM/RL Datasets} For the purpose of conducting RLHF-IME, we constructed two groups of datasets, one of which is the reward model training dataset group and the other is the prompt dataset group used in the reinforcement learning phase. 


 
All data are derived from the user behavior of real users recruited via the Internet to participate in the user improvement program. These users only need to choose the most satisfactory one among the given multiple model-generated candidate results as ordinary users do and do not need to do special processing for the rest bad results, not to mention the need to rank all the answers as the labelers do in \citet{ouyang2022training}. In this way, the consistency of the labeled data with the real input method application ensures the reliability of the data. At the same time, the simplicity of this labeling method makes the efficiency much higher than that of ranking labeling with specially trained labelers, which brings lower time and money costs.

As described in Section \ref{sec:imerlhfmethod}, for both the ranking and the binary classification annotation systems, we give labels to the samples collected in a fully automated way. Since the original data comes from a large number of real human users, we believe that the models trained based on it can highly fit human preferences and thus can play a positive role in guiding GeneInput in RLHF-IME. 




Regarding the ranking system, we would like to use the statistical probability of answer selection in a large user group to fit the human preference for a certain model-generated answer and use it as the basis for ranking. However, seeing that the number of users participating in the user improvement program for Conversational Assistance is relatively small, the answer ranking constructed based on this cannot fit the real human preference with high reliability, and then the reward model trained on this will be difficult to lead the GeneInput model to optimize in the right direction. Therefore, for Conversational Assistance, we only constructed datasets for the reward model with the binary classification system.

Please refer to Table \ref{tb:xfdatasetstatistics} for information on the data statistics of each dataset.

\begin{table}[t]
\small
\centering
\setlength\tabcolsep{2pt}
\begin{tabular}{lccccc}
\toprule
\multirow{2}*{\textbf{}} & \multicolumn{2}{c}{\textbf{IntelAssoc}} & \textbf{ConvAssist} & \multicolumn{2}{c}{\textbf{FK2C}} \\ 
 & Binary & Rank & Binary & Binary & Rank \\ \midrule

\scriptsize\textit{{SFT}} & & & & & \\ 
\textbf{Train Set} & \multicolumn{2}{c}{8M} & 6M & \multicolumn{2}{c}{12M} \\
\textbf{Validation Set} & \multicolumn{2}{c}{100K} & 100K & \multicolumn{2}{c}{100K} \\
\textbf{Test Set} & \multicolumn{2}{c}{2K} & 2K & \multicolumn{2}{c}{57K} \\

\midrule
\scriptsize\textit{{RM}} & & & & & \\ 
\textbf{Train Set} & 6.5M & 2.5M & 4.9M & 8.4M & 9.4M \\
\textbf{Test Set} & 1.6M & 0.6M & 1.2M & 2.1M & 2.3M \\

\midrule
\scriptsize\textit{{RL}} & & & & & \\ 
\textbf{Prompt Set} & \multicolumn{2}{c}{4.1M} & 1.9M & \multicolumn{2}{c}{4.0M} \\ 
\bottomrule
\end{tabular}
\caption{\label{tb:xfdatasetstatistics} Statistics of XF datasets.}
\end{table}

\subsubsection{Evaluation Metrics}




Due to the lack of objective evaluation criteria for Intelligent Association and Conversational Assistance, we used manual subjective metric - MOS. where two of each test case were generated by each model, and the generated results were independently scored by ten people, respectively, with scoring grades ranging from 1 (worst) to 5 (best), and the combined average score on all the test cases was the final score of the model on the task.



We use the precision of top-K (P@K) as the evaluation metric for the K2C task, which is often used in the past P2C tasks \cite{tanetal2022exploring}\cite{zhang-etal-2019-open}, indicating whether the desired result is included in the generated top-K results. Since the main focus in the input method is on the first and first screen results, we evaluate top1 and top5.




There are multiple training methods for reward models based on two annotation systems, therefore, we designed the following evaluation metrics for better evaluating the performance of models trained in each system.

    
\textbf{Accuracy-Rank} Accuracy-Rank($Acc_R$) is designed to evaluate the performance of reward models trained by the training method under the ranking annotation system. Its core idea is to compare how well the model's predicted scores match the ranked labeling information as displayed as Formula \ref{eq:remaccuracy}, where $label_{i}$ and $score_{i}$ represent the label value and the score given by the reward model for the $i_{th}$ sample.

    \begin{equation}
    \small
    \begin{aligned}
    Acc_R = \frac{\sum_{i=1}^{n}\sum_{j=1}^{n}\mathbb{I}_{label_{i} > label_{j}}\mathbb{I}_{score_{i} > score_{j}}\cdot1}{\sum_{i=1}^{n}\sum_{j=1}^{n}\mathbb{I}_{label_{i} > label_{j}}\cdot1}
    \label{eq:remaccuracy}
    \end{aligned}
    \end{equation}


\textbf{Accuracy-Binary} Accuracy-Binary($Acc_B$) is designed for evaluating the performance of reward models trained by the training methods under the binary classification annotation system. Positive and negative samples are paired and we compare the scores given by the reward models, and then calculate the percentage of pairs of samples for which the positive class wins. Calculated with reference to Equation \ref{eq:bemaccuracy}, where $N_{pos}$ and $N_{neg}$ denote the number of positive and negative samples respectively, and $score(pair_{i}^{pos})$ denotes the model's prediction value for the positive sample in the $i_{th}$ sample pair.

    \begin{equation}
    \small
    \begin{aligned}
    Acc_B = \frac{\sum_{i=1}^{N_{pos}\cdot N_{neg}}score(pair_{i}^{pos})>score(pair_{i}^{neg})}{N_{pos}\cdot N_{neg}}
    \label{eq:bemaccuracy}
    \end{aligned}
    \end{equation}

\subsubsection{Base Models}





In order to balance the model capability on various downstream tasks and the cost of hundreds of millions of calls per day in the Input Method scenario, all the experiments in this paper are based on the 2.6B version of iFlytek's self-developed LLM - Spark\footnote{Spark official website: \url{https://xinghuo.xfyun.cn/}}(except for the reward model). The model has a GPT-like structure containing 32 layers of transformers and is equipped with strong text generation capability after pre-training with a large amount of Chinese corpus.


For the reward model, we use the Chinese version of DeBERTa-v2-large \cite{fengshenbang} as the foundation model, and then add nonlinear layers, dense layers, etc. on top of it so that it generates the final model scores for the samples.

\subsubsection{Baselines}





\textbf{K2C} Since existing research on input methods mainly focuses on solving the 26-key prefect pinyin input, we compare with the following baselines based on the available PD dataset and TP dataset.

\begin{itemize}
\item GoogleIME is a commercial Chinese IME, which provides a debuggable API.
\item On-OMWA \cite{2017Tracing} is an adaptive learning model for new words in Chinese IME online word acquisition.
\item On-P2C \cite{zhang-etal-2019-open} is a neural Pinyin-Chinese conversion model, which enhances the model by online updating words to support open vocabulary learning.
\item Pinyin-GPT \cite{tanetal2022exploring} is a model that utilizes GPT by incorporating Pinyin as the previous context for Pinyin-Chinese translation.
\end{itemize}



\textbf{LLM} This paper selects chatGPT and GPT4 to explore the baseline performance of the current best large language models in input method tasks. We design more than 10 prompts for Intelligent Association, Conversational Assistance and K2C tasks in input methods respectively, and select the prompt with the best performance as the final prompt for testing on the test set.

\subsubsection{Configurations}



The SFT model is trained on 8 NVIDIA A100-80G for 1 week, the batch size is 128, we use a cosine annealing learning schedule with an initial learning rate of 1.6e-5, and we use the Adam optimizer with parameters of 0.9 and 0.95. The hyperparameters $\lambda$ and $\alpha$ are set as 1 and 0.5, respectively.



In RLHF-IME, the Spark is trained on the same devices as the SFT model for from 1 to 5 epochs where each epoch consists of 2 episodes and the batch size is 4096. We employ the AdamW optimizer \cite{adamw} with a peak learning rate of 9e-5 and a 10\% warm-up cosine scheduler. For reward modeling, we run experiments on 4 GPUs with smaller batch sizes (64 or 128) and learning rates (from 5e-6 to 1e-5) for different tasks. 



\subsection{Results}
\subsubsection{K2C Results}


\begin{table}[t]
\centering
\begin{tabular}{lcccc}
\toprule
 & \multicolumn{2}{c}{\textbf{PD}} & \multicolumn{2}{c}{\textbf{TP}} \\
\textbf{system} & P@1 & P@5 & P@1 & P@5  \\ \midrule

Google IME & 70.9 & 78.3  & 57.5 & 63.8 \\
On-OMWA & 64.6 & 72.9  & 57.1 & 71.1 \\
On-P2C & 71.3 & 80.5  & 71.9 & 89.7 \\
Pinyin-GPT & 73.2 & 84.1  & - & - \\ \midrule 

GeneInput & \textbf{88.4} & \textbf{96.2}  & \textbf{77.0} & \textbf{92.9} \\
\hspace*{3mm} - align & 88.1 & 95.9  & 76.4 & 92.5 \\ 
\hspace*{3mm} - align - pyseg & 82.1 & 92.4  & 70.1 & 88.6 \\  

\bottomrule
\end{tabular}
\caption{\label{tb:opendecoderesults}The results of the comparison between different methods on the PD and TP datasets.} 
\end{table}


\textbf{Results compared with existing methods} In Table \ref{tb:opendecoderesults}, the upper part of the results is the above comparison baseline effect, and the above effects are directly extracted from the Pinyin-GPT \cite{tanetal2022exploring} and On-P2C \cite{zhang-etal-2019-open} papers. The following shows our GeneInput method's effects on both datasets separately. It can be seen that our method, on both datasets, significantly surpasses the previous optimal effect in top1 and top5 metrics. This result demonstrates that while we have achieved full-mode K2C, we have not sacrificed single-mode effects, and compared to existing single-mode modeling methods, we still have a significant advantage, which reflects the effectiveness of our model.








\begin{table}[t]
\centering
\begin{tabular}{lcccc}
\toprule
 & \multicolumn{2}{c}{\textbf{26-key}} & \multicolumn{2}{c}{\textbf{9-key}} \\
\textbf{system} & P@1 & P@5 & P@1 & P@5  \\ \midrule

\multicolumn{2}{l}{\small\textit{{Perfect Pinyin}}} & & & \\
Google IME & 88.0 & 90.1  & 75.3 & 77.1 \\
GeneInput & \textbf{94.2} & \textbf{99.5} & \textbf{92.0} & \textbf{98.4} \\ \midrule 

\multicolumn{2}{l}{\small\textit{{Abbreviated Pinyin}}} & & & \\
Google IME & 30.2 & 32.2  & \textbf{2.4} & 3.3 \\
GeneInput & \textbf{67.0} & \textbf{86.7}  & 1.6 & \textbf{4.6} \\ \midrule  

\multicolumn{2}{l}{\small\textit{{Random Abbreviated Pinyin}}} & & & \\
Google IME & 65.1 & 66.9  & 41.3 & 43.4 \\
GeneInput & \textbf{81.5} & \textbf{95.3}  & \textbf{73.4} & \textbf{88.4} \\ \midrule  

\multicolumn{2}{l}{\small\textit{{Pinyin with Noise}}}  & & & \\
Google IME & 55.2 & 67.2  & 7.2 & 11.3 \\
GeneInput & \textbf{75.2} & \textbf{90.7}  & \textbf{46.2} & \textbf{67.5} \\ 

\bottomrule
\end{tabular}
\caption{\label{tb:xfdecoderesults}Results of different input modes on XF dataset.} 
\end{table}










\textbf{Results on Full-mode K2C} Due to the current lack of research on modeling other input modes and full-mode, we compare with the open commercial Chinese input method Google IME on our self-built XF dataset to verify the effect of the full-mode K2C model.

In Table \ref{tb:xfdecoderesults}, our method has a significant improvement over googleIME on each input mode test set except for the one in 9-key abbreviated pinyin input mode where it performs worse than GoogleIME. Especially on the input data to be noised, we also have a good performance. And for 9-key input, users usually do not choose to perform abbreviated pinyin input, so we think its effect on the 9-key abbreviated pinyin test set is unimportant. From the above results, we can see that we have achieved a good result in each input mode, which shows that our full-mode unified modeling is successful.


\subsubsection{RLHF-IME Results}





\begin{table}[t]
\centering
\setlength\tabcolsep{4pt}
\begin{tabular}{lccc}
\toprule
\textbf{Method} & \textbf{IntelAssoc} & \textbf{ConvAssist} & \textbf{FK2C} \\ \midrule

\small\textit{{Rank}} & & & \\ 
Query-Wise & 68.5 & - & 73.5 \\ 
\midrule 

\small\textit{{Binary}} & & & \\ 
Sample-Wise & 99.3 & 77.7 & 98.9  \\
Class-Wise  & 93.0 & 73.1 & 81.2 \\
Contra-Wise & 97.7 & 67.9 & 97.9 \\
Batch-Wise & \textbf{99.5} & \textbf{78.1} & \textbf{99.6}  \\  

\bottomrule
\end{tabular}
\caption{\label{tb:rmresults}Results of all kinds of training methods for reward modeling.} 
\end{table}


\textbf{RM Results} Table \ref{tb:rmresults} shows the results of reward models trained based on each training method in Intelligent Association, Conversational Assistance and Full-mode K2C. In the ranking system, there is only one training method so comparison between different methods is not possible. Hence we tested the model in different testing scenarios similar to those in Table \ref{tb:rmscoreexamples} and the results indicated that the models perform highly consistently with human preference. As explained in section \ref{sec:xfdatasets}, there is no dataset constructed in the ranking system for Conversational Assistance, so the corresponding result is missing. In the binary classification system, Batch-Wise achieves the best results on all tasks, and interestingly, Sample-Wise conducting simple binary classification learning is observed to have the smallest gap (within 0.5 points) with Batch-Wise. Contra-Wise also achieves good results on Intelligent Association and Decode with the supervised comparison learning strategy. However, there is a fact that the texts in the input method scenario are flexible in context and short in length, which leads to a blurring of the boundaries of learning, so there is still a gap of about 3 or 4 points with Batch-Wise. Class-Wise performs the worst in line with the design expectation because it is the method with the coarsest learning granularity, and it is difficult for the model to capture fine-grained sample-level preference characteristics when learning the positive and negative classes as a whole to capture the inter-category differences.


\begin{table}[t]
\centering
\setlength\tabcolsep{3pt}
\begin{tabular}{lccc}
\toprule
\small{\textbf{Model}} & \small{\textbf{IntelAssoc}} & \small{\textbf{ConvAssist}} & \small{\textbf{FK2C}} \\ \midrule
\footnotesize{\textit{public}} &  &  & \\
ChatGPT & 3.88 & 4.26 & 12.3  \\ 
GPT-4 & 4.41 &4.35 & 18.1  \\ \midrule
\footnotesize{\textit{GeneInput}} &  &  & \\
Spark &  &  & \\
\hspace*{1mm} + \small{SFT} & 4.38 & 4.25 & 81.0 \\
\hspace*{3mm} + \small{RLHF-IME rank} & 4.40 & - & 82.8 \\ 
\hspace*{3mm} + \small{RLHF-IME binary} & \textbf{4.43} & \textbf{4.52} & \textbf{84.6} \\ 
\bottomrule
\end{tabular}
\caption{\label{tb:sftrlhfresults} Results of LLMs on IntelAssoc, ConvAssist and FK2C.}
\end{table}


\textbf{SFT/RL Results} The results of how large language models perform on Intelligent Association, Conversational Assistance and Full-mode K2C are provided in Table \ref{tb:sftrlhfresults}. As the base model Spark is pre-trained on large-scale unlabeled data and has not been optimized with instruction tuning, the results of it on these tasks in the input method scenario are not provided. Since ChatGPT and GPT-4 are not capable of understanding the relations between 9-key input sequences and corresponding Chinese sequences, we only show the average results on the 26-key test set of XF dataset for comparing the performance of input sequence decoding among LLMs, for more detailed results of GeneInput on FK2C please refer to Table \ref{tb:opendecoderesults} and Table \ref{tb:xfdecoderesults}. Apparently, the existing LLMs for general purpose perform extremely poorly on FK2C and cannot meet the productization requirements in the input method scenario, so it is necessary to propose our input-method-specific LLM. On Intelligent Association and Conversational Assistance, after SFT, Spark has been equipped with competitive capabilities. However, there is still a gap with the state-of-the-art LLMs, and especially on Conversational Assistance, Spark with SFT does not even catch up with ChatGPT. Fortunately, after optimized with RLHF-IME, Spark outperforms GPT-4 on both tasks, especially on Conversational Assistance where the MOS score is even higher by 0.17. In addition, by comparing the models with RLHF-IME based on reward models trained from different annotation systems on Intelligent Association and Full-mode K2C, we arrive at the conclusion that it is more effective to use the binary classification system to train RMs instead of the ranking system in the input method scenario.

\subsection{Ablation Experiment}


In this section, we do some ablation experiments to help understand the role of increasing pinyin segmentation and decoding alignment constraints. Since our alignment constraints are bridged by intermediate processes of pinyin segmentation, we first remove the alignment constraints and then remove the segmentation process in turn. As shown in the lower part of Table \ref{tb:opendecoderesults}, after removing the alignment constraints, top1 has a loss of 0.5 points, but after removing pyseg, top1 has a drop of about 6 points on both datasets. This result fully reflects the important role of adding intermediate processes of pyseg class into output expansion during training. Of course, alignment constraints also play a certain role, but since the model already has a good alignment relationship after adding pyseg, it significantly reduces the room for improvement of explicit alignment constraints.

\subsection{Case Study}




\subsubsection{Personalization} 
For the personalization of the input method, we have done some simple case analysis. As shown in Table \ref{tb:IMEpersonalexamples}, for the same user inputs, we add different input expansions and can get different output results. For example, in the intelligent association task, for the same input “I don’t have time tonight”， we add different user information to the input expansion, and then we can output different results that are more in line with the user’s occupational characteristics, thus achieving personalized output. Similarly, in the FK2C task, according to different context information or user feature descriptions, corresponding more accurate results can be output. We leave a more detailed analysis as future work.



\subsubsection{Reward Modeling} As mentioned in Section \ref{sec:imerlhfmethod}, the reward model determines the upper bound of RLHF performance to some extent. Therefore, we expect the model scores on different answers highly compatible with human preferences. Table \ref{tb:rmscoreexamples} shows examples of our reward model scoring for Conversational Assistance in different test scenarios. It is clear that it has human-consistent perceptions of ordinary answers and those with high quality, shows significant resistance to profanity and irrelevant replies, and at the same time has good discriminative ability for answers to different generative tasks (e.g., the intelligent association answer in the table). Consequently, we firmly believe that it can give good feedback to the large language models for IME in line with human preferences during reinforcement learning.

\begin{table*}
\small
\centering
\setlength\tabcolsep{4pt}
\begin{tabular}{lccc}
\toprule
\textbf{Task Type} & \textbf{Input} & \textbf{Input Extend} & \textbf{Output Top1} \\ \midrule
 &  & - & \makecell[c]{明天可以吗\\ \small How about tomorrow?}  \\
IntelAssoc & \makecell[c]{今晚没时间\\ \small There's no time tonight.} & \makecell[c]{用户：学生\\ \small User: Student} & \makecell[c]{作业还没写完呢\\ \small I haven't finished my homework yet} \\ 
 &  & \makecell[c]{用户：程序员\\ \small User: Programmer} & \makecell[c]{晚上还得加班呢\\ \small I have to work late at night}  \\ \midrule
 &  & - & \makecell[c]{万一\\ \small in case}  \\
FK2C & \makecell[c]{wany } & \makecell[c]{上文：来找我\\ \small Context: Come to me} & \makecell[c]{玩呀\\ \small playing } \\ 
 &  & \makecell[c]{用户：习惯全简拼输入\\ \small User: Used to abbreviate pinyin input} & \makecell[c]{我爱你呀\\ \small I love you}  \\
  &  & \makecell[c]{用户词：婉莹\\ \small User word: Wan Ying} & \makecell[c]{婉莹\\ \small Wan Ying}  \\
\bottomrule
\end{tabular}
\caption{\label{tb:IMEpersonalexamples} Examples of IME personalization (the corresponding translations in English are provided below the sentences in Chinese).}
\end{table*}

\begin{table*}
\small
\centering
\setlength\tabcolsep{4pt}
\begin{tabular}{lccc}
\toprule
\textbf{Test Type} & \textbf{Query} & \textbf{Answer} & \textbf{RM Score} \\ \midrule
Good Conversational Assistance &  & \makecell[c]{宝贝早安喔，希望你今天一天都开开心心\\ \small Good morning, my love. I hope you are happy all the day.} & 0.974  \\
Bad Conversational Assistance &  & \makecell[c]{嘿嘿，早上好哦\\ \small Hey, hey, good morning. (fondly)}
 & 0.941 \\ 
Bad Language & \makecell[c]{早上好\\ \small Good morning.} & \makecell[c]{去你妈的\\ \small Fuck you, man.} & 0.022  \\ 
Irrelevant Words &  & \makecell[c]{你还爱他吗\\ \small Do you still love him?} & 0.083  \\ 
Intelligent Association &  & \makecell[c]{昨晚睡得怎么样\\ \small How did you sleep last night?} & 0.139  \\ 
\bottomrule
\end{tabular}
\caption{\label{tb:rmscoreexamples} Examples of RM for Conversational Assistance scoring in different test settings with the same query (the corresponding translations in English are provided below the sentences in Chinese).}
\end{table*}

\section{Conclusion}
In this work, we explore how the next-generation generative paradigm \textbf{GeneInput} can be employed to uniformly model typical tasks within IMEs using generative models through prompts. We harness the text generation capability of the model to empower input method, extending their functionality beyond mere P2C and full-mode K2C is realized for the first time. Furthermore, we introduced four novel reward-model training methods based on user feedback, allowing online model updates without the need for external annotated data, and resulting in state-of-the-art performance across all tasks. In the future, we plan to address more auxiliary input functions and further reduce the model size to make it capable of running efficiently on most smartphones while maintaining high performance.


\bibliography{anthology,custom}
\bibliographystyle{acl_natbib}

\appendix


\end{CJK*}
\end{document}